\documentclass[%
reprint,
 amsmath,amssymb,
 aps,
pre,
]{revtex4-2}

\usepackage{graphicx}%
\usepackage{dcolumn}%
\usepackage{bm}%
\usepackage{mathrsfs}
\usepackage{wrapfig}
\usepackage{makecell}

\begin{document}

\title{Variational Continuation for Double Pendulum Periodic Orbits}%

\author{Leo Yao}
\email{leoy@mit.edu}
\affiliation{%
 Department of Physics, Massachusetts Institute of Technology, Cambridge, Massachusetts
}%
\affiliation{%
 The NSF AI Institute for Artificial Intelligence and Fundamental Interactions
}%
\author{Ziming Liu}%
\email{zmliu@mit.edu}
\affiliation{%
 Department of Physics, Massachusetts Institute of Technology, Cambridge, Massachusetts
}%
\affiliation{%
 The NSF AI Institute for Artificial Intelligence and Fundamental Interactions
}%
\author{Max Tegmark}%
\email{tegmark@mit.edu}
\affiliation{%
 Department of Physics, Massachusetts Institute of Technology, Cambridge, Massachusetts
}%
\affiliation{%
 The NSF AI Institute for Artificial Intelligence and Fundamental Interactions
}%

\date{\today}%

\begin{abstract}
We present a Hessian-based approach to numerically continue periodic orbits in dynamical systems. A loop (periodic orbit candidate) is parametrized as a Fourier series; a loss function is defined based on the deviation of the loop from the physical differential equations. Unlike previous work relying on hand-derived Jacobians, our method automates the process by leveraging automatic differentiation, a common machine learning technique. The continuation direction can be determined by the flat directions of the loss landscapes (directions with zero eigenvalues), making the search of periodic orbits efficient and guided. Our method is integrator-free, precisely initializes oscillations around unstable fixed points, and efficiently detects orbit family intersections and subharmonic bifurcations. As a demonstration, we present full continuations of periodic double pendulum oscillations from fixed points, showing bifurcations along orbit families and categorizing branches of periodic orbits. In particular, we find periodic orbits where both pendulum masses are never simultaneously at rest, which to our knowledge has been missing in the literature.
\end{abstract}

\maketitle

\section{Introduction}\label{introduction.sec}

Finding and analyzing periodic orbits plays a key role in understanding dynamical systems \cite{Gutzwiller1990,Gelfreich_2002,Sbano2011}. Periodic orbits can not only facilitate calculations of statistical properties of trajectories (e.g., via cycle expansions)~\cite{Artuso_1990,Lan_2010,PhysRevE.102.052220} but also are important due to the inherent simplicity of recurring trajectories. Through both statistical methods and intuition, periodic orbits offer a window through which we can start to understand the dynamics of otherwise difficult-to-interpret physical systems. 

Although various numerical methods have been developed to search for periodic orbits~\cite{npg-14-615-2007,FARANTOS199591,PhysRevE.53.1206,PhysRevE.64.026214}, there does not exist an efficient method to map out the full orbit space. Our contribution is a continuation method that leverages automatic differentiation and Hessian eigendecompositions to efficiently navigate through the space of periodic orbit solutions.

\begin{figure*}[t]
  \centering
  \includegraphics[width=\textwidth]{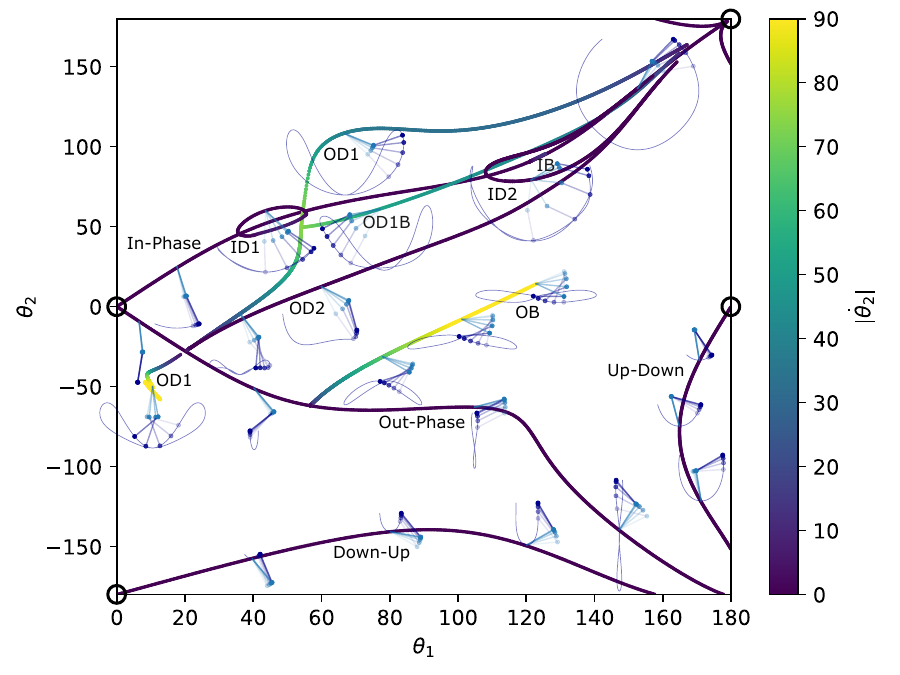}
  \caption{Converged double pendulum periodic orbits in the Poincaré section with $\dot{\theta}_{1}=0$; bifurcating branches intersect and extend off continuous orbit families between fixed points. Fixed points circled in black. \textit{In-Phase}, \textit{Out-Phase}, \textit{Down-Up}, and \textit{Up-Down} start from fixed points and run to period divergences. Bifurcation branches abbreviated with \textit{I} and \textit{O} for base branch, \textit{B} for branching at same period, \textit{D} for period-doubling, and numbered for uniqueness.}
  \label{full-plot.figure}
\end{figure*}

We use the double pendulum to demonstrate our method, as shown in Figure~\ref{full-plot.figure}. Although the double pendulum is one of the most elementary systems in physics, to the best of our knowledge, we are the first to survey bifurcations in the periodic orbit manifold and visualize them on a Poincaré section. This diagram not only shows bifurcation structures, but also reveals more complicated orbits, particularly those without a moment when both masses are at rest. This simple fact, albeit easily understood, has been missing from the literature. 

We briefly review the two basic numerical methods for finding periodic orbits. The shooting method first extracts close recurrences from integrated initial conditions and then converges a periodic orbit via Newton-Raphson descent or other methods~\cite{MESTEL1987172}. However, a fundamental property of chaotic systems is the exponential divergence of close trajectories, requiring an increasingly precise initial guess for longer orbits or higher-dimensional systems, even close to fixed points. Converging longer orbits becomes exponentially more difficult, or even impossible due to accumulating numerical precision errors. Variations, such as multiple shooting~\cite{S_NCHEZ_2010}, alleviate the convergence issue partially but not fully, and introduce additional complexities of their own.

Another orbit-finding method that avoids convergence issues is the variational method. Instead of varying initial conditions and evolving until obtaining a closed orbit, the initialization is a closed loop that does not necessarily satisfy the equations of motion. The loop is then varied until satisfying the evolution equations, resulting in a periodic orbit. As the loop is adapted to the equations of motion locally, no integration of conditions and subsequent exponential divergences occurs. Variations on this method include choice of loop representation, method of optimization, and method of loop initialization. When introduced by Lan and Cvitanović, the orbit was represented as evenly spaced points in time, with the first derivative given by a four-point approximation~\cite{PhysRevE.69.016217}. Representations with different spacings, such as a weighting between phase space and time spacing~\cite{PhysRevE.98.042204}, and representations involving Fourier coefficients in time have also been implemented since. Methods of optimization range from Newton descent and other Jacobian-based methods to adjoint-free and other linear algebra methods~\cite{PhysRevE.105.014217}.

In most periodic orbit studies, initializations are made either by perturbing from known fixed points or by extracting close recurrences from a grid of initial conditions. These methods do not incorporate the underlying orbit structure into additional initializations as it is probed, finding solutions first and classifying families of solutions later with human input. Numerical continuation methods have been previously applied in periodic orbit searches, but the majority of attempts have used older integrator methods. Computing the tangent vector in pseudo-arclength continuation has involved integrating the Jacobian~\cite{Wulff_2006}, a computationally expensive and numerically unstable task for longer orbits, and there has not been recent development on other propagation techniques. It is our motivation to revisit this continuation problem, given great recent advancement of computational tools, especially the technique of automatic differentiation. 

The ability to map out the whole periodic orbit spectrum can shed great light on the properties of dynamical systems. The two simplest physical systems demonstrating chaotic behavior and a rich periodic orbit spectrum are the double pendulum and plane-circular restricted three-body problem (PCR3BP). Two Hamiltonian systems each with a four-dimensional state space, both have multiple known fixed points, stable and unstable, and small perturbations from them forming periodic orbits. From periodic orbit family theory~\cite{Henon_1997,Wintner_1932,Birkhoff1936,BARANGER198895}, we expect periodic orbits to be organized into continuous, single-parameter families. An individual family of orbits cannot bifurcate or arbitrarily conclude; orbit families can only terminate at fixed points, or if the period or a phase space coordinate diverges. Bifurcations in the orbit spectrum are caused by crossings of families at the same and higher multiples of periods, and these crossings also correspond to changes in orbit stability. These connected, one-dimensional families of orbits make dimension-four Hamiltonian systems the perfect application for numerical continuation techniques.

\begin{figure}[t]
  \centering
  \includegraphics[width=0.8\columnwidth]{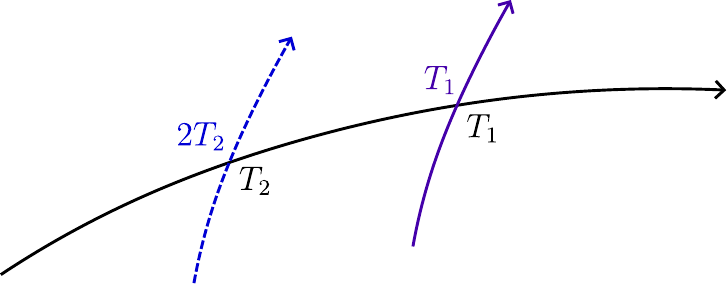}
  \caption{An example of a periodic orbit spectrum consisting of multiple orbit families. At period $T_{1}$, a crossing orbit family creates a bifurcation in the orbit spectrum, shown in purple. At period $T_{2}$, a period-doubling bifurcation occurs due to a crossing of an orbit family with twice the period, shown in dashed blue.}
\end{figure}

While orbits in the PCR3BP have been extensively studied, most work on the double pendulum has focused on systems with external driving and damping forces. These are studied to model various real-world applications, but lead to non-Hamiltonian systems with perturbed steady states and limit cycles~\cite{Yu_1998,Yagasaki_1996,Liu_2023,Williams_2023}. For the pure double pendulum, fixed points (with the pendulum bobs vertically up or down) and small angle periodic orbits are known, as well as basic numerical investigations, such as the Lyapunov exponent and autocorrelation function~\cite{Calv_o_2015,Jim_nez_L_pez_2024}. Periodic orbits are occasionally found as a byproduct of other explorations, such as auto-discovering conserved quantities~\cite{PhysRevLett.126.180604}, but no systematic search has been conducted. More recently, there has been some research on homoclinic and heteroclinic orbits, which has also yielded orbits around unstable fixed points~\cite{Kaheman_2023}.
However, a full numerical survey of double pendulum periodic orbits has not been conducted in the literature. Orbits are theoretically predicted to form continuous families, but extensions of small angle oscillations to higher amplitudes has not yet been attempted. While individual examples of more complicated periodic orbits exist, no methodical investigations of the orbit spectrum and bifurcations leading to non-stationary periodic orbits have been conducted.

In this paper, we introduce a novel Hessian-based variational continuation method for periodic orbits. In Section~\ref{method.sec}, we represent a periodic orbit loop under a Fourier parametrization and study Hessian eigenvectors of the variational loss. In Section~\ref{results.sec}, we demonstrate the automatic discovery of small oscillations around unstable fixed points, continuation of orbit families with precise loop initializations, and detection of bifurcations in the orbit spectrum. We present full extensions of symmetric periodic orbits from small oscillations to period divergence, examples of orbit family crossings and spectrum bifurcations, and discuss integrating the Hessian method with other orbit-finding techniques. We summarize our conclusions in Section~\ref{conclusion.sec}.

\section{Method}\label{method.sec}

\subsection{Variational Orbit Setup}

We consider a general dynamical system parametrized by some state vector $\mathbf{z}$ in a phase space $\mathcal{M}$, where time evolution $\mathbf{z}(t)$ is determined by the ordinary differential equation:
\begin{equation}
  \frac{\mathrm{d} \mathbf{z}}{\mathrm{d} t} = \mathbf{f}(\mathbf{z}) \label{systemdiffeq}
\end{equation}
A periodic orbit is defined by an initial condition $\mathbf{z}_{0}$ in phase space and a time $T > 0$ such that $\mathbf{z}(T)= \mathbf{z}_{0}$.

To set up our variational method, we take a closed loop $\mathcal{L}$, defined as a period $T$ and a trajectory $\mathbf{z}(t)$ in phase space, with $0 \leq t < T$, $\mathbf{z} \in \mathcal{M}$, and $\mathbf{z}(T)= \mathbf{z}(0)$. At each point $\mathbf{z}(t)$ on the trajectory, we can quantify the deviation of the loop trajectory from physical evolution by taking the squared norm of the difference:
\begin{equation}
  \ell(t) = \left| \left| \frac{ \mathrm{d} \mathbf{z}(t) }{ \mathrm{d} t } - \mathbf{f}(\mathbf{z}(t))  \right| \right|^{2} \label{pointnorm}
\end{equation}
We then average over the entire trajectory to obtain a loss function for the loop:
\begin{equation}
  \ell(\mathcal{L}) = \frac{1}{T} \int_{0}^{T} \ell(t) \, \mathrm{d}t = \frac{1}{T}\int_{0}^{T} \left| \left| \frac{ \mathrm{d} \mathbf{z}(t) }{ \mathrm{d} t } - \mathbf{f}(\mathbf{z}(t)) \right| \right|^{2} \, \mathrm{d}t \label{looploss}
\end{equation}
This loss function quantifies the overall deviation of the loop from a physical trajectory. Although the loop space is in principle infinite-dimensional, we need to choose a finite-dimensional parameterization. We will parameterize the loop as a Fourier series (as detailed in Section~\ref{implementation.subsec}), but for now, it suffices to denote the parameters as a vector $\bm{\theta}$, so $\ell(\mathcal{L})$ becomes $\ell(\bm{\theta})$. Zeroes of $\ell$ correspond to loops that always locally match physical evolution, and therefore are periodic orbits. As the loss function is always nonnegative, at periodic orbits the loss $\ell$ and gradients $\nabla_{\bm{\theta}} \ell$ both vanish. 

\begin{figure}[ht]
  \centering
  \includegraphics[width=.6\columnwidth]{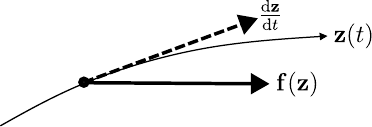}
  \caption{For an arbitrary trajectory $\mathbf{z}(t)$ in phase space, the loop tangent $\frac{ \mathrm{d} \mathbf{z}(t) }{ \mathrm{d} t }$ may not necessarily match the physical evolution $\mathbf{f}(\mathbf{z}(t))$. For a periodic orbit, these vectors must match across the entire loop.}
\end{figure}

\subsection{Hessian Analysis}

Once we have a single periodic orbit, how can we use it to obtain a new nearby orbit? Traditional continuation methods start from an initial condition, then integrate the entire Jacobian matrix~\cite{Net_2015}. This integration is both numerically unstable and ignores the originally discovered orbit structure by just using an initial condition. By instead considering the entire orbit loop and leveraging automatic differentiation, we can more easily obtain such a perturbation direction by analyzing the Hessian of the variational loss landscape.

The Hessian $\mathbf{H}$ of the loss function $\ell$ describes the local curvature of the loss around a parametrized loop $\mathbf{\theta}$ within a parameter space. For a periodic orbit, $l$ is at a local minimum, so the Hessian $\mathbf{H}(\mathbf{\theta})$ must be positive semi-definite. The eigenvalues provide information about curvature magnitudes and associated directions.

For a zero eigenvalue, the associated eigenvector gives a direction in which perturbing $\mathbf{\theta}$ maintains zero loss, and therefore maintains a periodic orbit. For a nontrivial orbit (not a fixed point), the full Hessian must have at least one such flat direction, corresponding to shifting the phase of the loop in time. Additional vanishing eigenvalues correspond to additional perturbation directions for the loop, creating a subspace of possible perturbations. The dimension of this nullspace, and therefore the dimensionality of the connected periodic orbit space at $\mathbf{\theta}$, is given by the number of zero eigenvalues of the Hessian $\mathbf{H}(\mathbf{\theta})$.

We can also consider a subspace of the loop parameter space, restricting the possible perturbations to the loop. Taking the subspace Hessian then gives only local curvatures corresponding to possible loop perturbations within this subspace. For example, enforcing a phase condition removes time evolution from the possible perturbations, removing the corresponding zero eigenvalue. By restricting trivial or known perturbations from the subspace until there is a unique zero eigenvalue, we can extract the direction of chosen adjacent connected orbits by reading off the remaining unique flat direction.

An adjacent orbit initialization can be made by taking a finite step in loop space along the flat direction. The orbit family may not be perfectly linear, but for a sufficiently small step size, the initialization will be close to the orbit family, and can be converged to a periodic orbit on the family. This directional step is the same tangent propagation step used in pseudo-arclength continuation, but made in loop space. Further, as the new initialization is an entire loop, the subsequent periodic orbit optimization benefits from the much larger radius of convergence of the variational method.

\subsection{Hessian Applications}

\begin{figure*}[t]
  \centering
  \includegraphics[width=.8\textwidth]{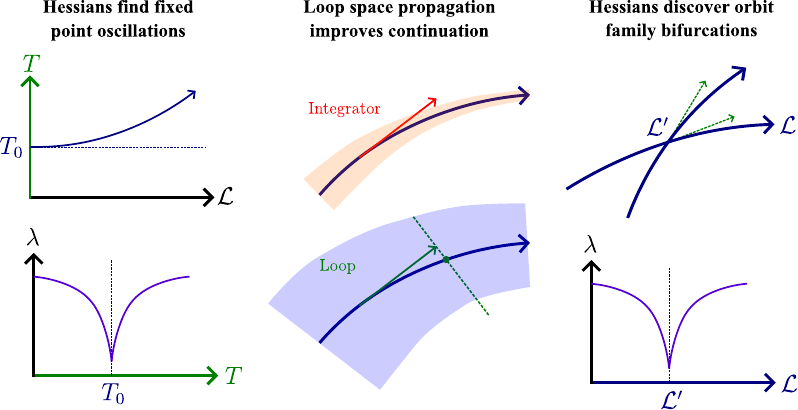}
  \caption{On left, a branch of orbits $\mathcal{L}$ emerging from a fixed point at period $T_{0}$ (above) is found by sweeping period $T$ at the fixed point, and detecting a drop in the minimal eigenvalue $\lambda$ (below). At center, the increased radius of convergence (shaded) of the variational method allows for exploration further along orbit families, with an orthogonal loop space constraint. On right, a crossing of orbit families at $\mathcal{L}'$ (above) is found by detecting a drop in the off-branch eigenvalue $\lambda$ (below).}
  \label{variational-diagrams.figure}
\end{figure*}

The method of Hessian analysis allows brute force methods for finding periodic orbits, such as grid search, to be replaced by a more methodical approach leveraging the underlying geometric structure of periodic solutions.

Fixed points are (trivial) periodic orbits for any period $T$, and the starting point for periodic orbit searches. While small amplitude periodic perturbations around stable fixed points can be found by linearization, orbits around unstable fixed points require manual converging, which can be increasingly difficult for chaotic system dynamics and may collapse back down to the fixed point.

The Hessian method allows for precise initializations of small amplitude periodic perturbations, even in highly chaotic systems. Loops can be initialized to the fixed point for various periods $T$, and the existence of periodic perturbations can be tested by computing the Hessian. A drop of the minimal eigenvalue to zero indicates a period $T$ with a periodic perturbation, and this perturbation is given by the corresponding eigenvector. By sweeping values of $T$, all possible small oscillations around fixed points, which are starting points for periodic orbit branches, can be systematically discovered.

Continuation of periodic orbit branches can be improved by the Hessian method. The larger radius of convergence in loop space allows for larger steps to be taken and longer orbits to be continued, allowing exploration further along orbit families even when integrator methods fail due to precision issues. The eigenvalue direction also gives an orthogonal orbit constraint in loop space, allowing for regular steps in pseudo-arclength to be taken while better reflecting the overall structure of the orbit compared to just an initial condition.

The Hessian method also offers a means to detect bifurcations of periodic orbit families. If an orbit family crosses another, there will be two unique flat directions at the crossing point, one along each family. This gives a two-dimensional nullspace at the bifurcation point, and can be detected as a second eigenvalue drops to zero as it is approached. Constraining the subspace to eliminate the known orbit family at the bifurcation point, the direction of the crossing orbit family can be determined and used to obtain an initialization for it.

Subharmonic bifurcations can be detected with a modification to the loop. For a period-doubling bifurcation, a new loop can be constructed with two windings of the original orbit. This doubly-wound loop can still be propagated along the original family, though with all components at doubled (even) frequencies. However, subharmonic (odd frequency) components can now also be perturbed, allowing a second flat direction corresponding to the period-multiplied bifurcation. By transforming a loop representation to a multiply-winded version and testing for zero eigenvalues, other subharmonic bifurcations in the orbit spectrum can also be detected and initialized.

\subsection{Implementation}\label{implementation.subsec}

We consider the ideal double pendulum with equal point masses and equal-length massless arms, setting all parameters to $1$ ($m_{1} = m_{2} = l_{1} = l_{2} = g = 1$). A system state is specified by arm angles $\theta_{1}, \theta_{2}$ and angular velocities $\dot{\theta}_{1}, \dot{\theta}_{2}$. A loop $\mathcal{L}$ is represented by a period $T$ and a closed trajectory $\mathbf{z}(t)$. We parametrize $\mathbf{z}(t)$ with a Fourier decomposition in time with a finite maximum frequency cutoff $K$:
\begin{align}
  \mathbf{z}(t) = \mathbf{a}_{0} + \sum_{k=1}^{K} \mathbf{a}_{k} \cos\left(\frac{2k\pi t}{T}\right)+\mathbf{b}_{k} \sin\left(\frac{2k\pi t}{T}\right), \notag \\ \text{where }  0\leq t\leq T.
\end{align}
We implement computations in PyTorch~\cite{NEURIPS2019_9015} with \texttt{float64} precision, leveraging autograd for Hessian computation. As parameters and gradients vary by orders of magnitude between low and high frequency components, we use the Rprop algorithm~\cite{riedmiller1993direct}, which adapts step sizes per parameter based solely on the signs of the gradients. We optimize to a minimum of the integrated error, and use a convergence condition of $\sqrt{ \ell } < 10^{-10}$.

Higher energy, longer period orbits require higher frequency components to be suitably represented by the Fourier parametrization. As we propagate along a family of orbits, we monitor the final converged value of $\ell$, increasing the frequency cutoff $K$ if the convergence condition is not reached. Symmetries in the orbit are also exploited to reduce the number of parameters optimized. For spatially symmetric orbits of the double pendulum, we evolve (require gradient on) only cosine positional components and sine momentum components, enforcing the symmetry constraint and halving the parameter count. This constraint both enforces a phase condition and avoids accidentally propagating to perturbed orbits. 

\section{Results}\label{results.sec}

\subsection{Fixed Point Analysis}

We start by using a sweep of period $T$ to analyze the fixed points of the double pendulum. We choose a step size of $\Delta T=0.01$, which controls the precision of the detected period $T$ and the accuracy of the corresponding orbit initialization. As the loop representations of fixed points can be immediately initialized (all non-constant Fourier components are zero), the only computation necessary is computing the Hessian, and the entire sweep takes a few minutes to run for $K=16$ Fourier terms.

Plotting the minimum Hessian eigenvalue against period $T$ gives sharp drops to zero at $T \approx 3.39$ and $T \approx 8.21$ for the stable fixed point (both masses down), and $T \approx 5.28$ around both unstable saddle points (one mass down, one mass up). These $T$ values match the theoretically known periods of oscillation for normal modes. No eigenvalue drop is noticed for the unstable fixed point (both masses up), as small-amplitude periodic orbits do not exist around it.

The corresponding eigenvectors give precise initializations for periodic orbits. Taking a step of size $\Delta \theta = 0.5^{\circ}$ in phase space, we can compute an initial condition from the loop $\mathcal{L}$ at time $t=0$, then integrate for a time $T$ given by our detected period. Comparing the final result of the integration with the starting point, we obtain relative deviations from the perfect periodicity of within $2 \%$ compared to the phase space step size (Table~\ref{fixed-point-errors.table}). Our method is robust to local stability, providing precise initial conditions around both stable and unstable fixed points without the need for a computationally expensive search.

\begin{table}[ht]
  \caption{For each of the detected normal modes, a step in loop space corresponding to a phase space deviation of $0.5^{\circ}$ is taken, an initial condition is obtained by evaluating the Fourier parametrization at $t=0$, and this phase space condition is integrated for the detected period $T$. The deviation of the integration result from the original condition is small compared to the step size of $0.5^{\circ}$, implying a precise initialization corresponding to a periodic orbit.}
  \begin{ruledtabular}
  \begin{tabular}{ccc}
  \thead{Normal Mode} & \thead{Integrated Deviation\\ (degrees)} \\ %
  \hline
  Down-Down, In-Phase & $1.76 \times 10^{-4}$ \\ %
  Down-Down, Out-Phase & $1.02 \times 10^{-3}$ \\ %
  Down-Up & $0.0982$ \\ %
  Up-Down & $0.0976$ \\ %
  \end{tabular}
  \end{ruledtabular}
  \label{fixed-point-errors.table}
\end{table}

\begin{figure}[ht]
  \centering
    \includegraphics[width=\columnwidth]{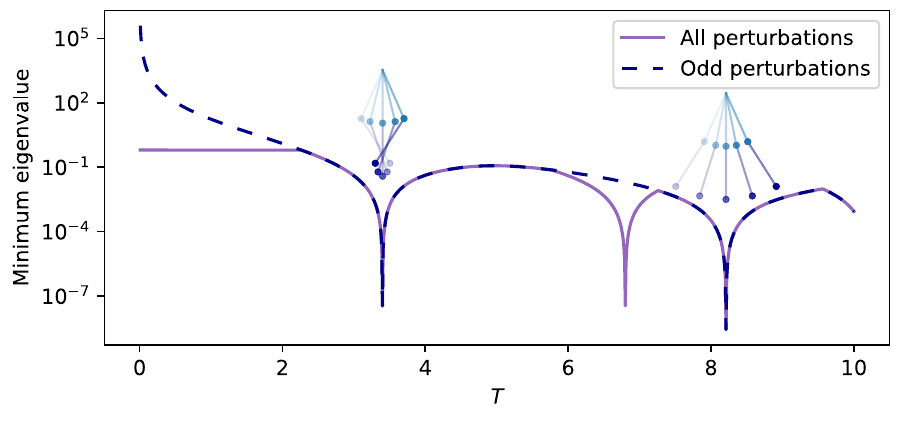}
    \caption{Minimum eigenvalue for the period $T$ with both masses down ($\theta_{1}=0$, $\theta_{2}=0$); eigenvalue minimums give out of phase and in-phase normal modes and correct corresponding periods $T$.}
    \label{down_down_evalues.figure} %
\end{figure}

If we take the Hessian of all free parameters for the stable fixed point, we also obtain an additional eigenvalue dip at $T \approx 6.78$. This corresponds to the same normal mode at the $T \approx 3.39$ eigenvalue zero, but with two oscillations in a single loop instead of one. Constraining the free parameters of the Hessian to require antisymmetry in the two halves of the orbit (only odd frequency components) removes this period-doubled version of the oscillation from the eigenvalue zeroes (Figure~\ref{down_down_evalues.figure}).

\subsection{Symmetric Orbit Propagation}

From each normal mode, we propagate a family of symmetric orbits by repeatedly converging an initialization to a periodic orbit, taking the Hessian, and taking a loop space step in the zero-eigenvalue direction. During convergence, we constrain the training dynamics perpendicular to the step direction, ensuring that propagation proceeds along the orbit family. We use a step size of $0.5^{\circ}$ for all orbit families. We constrain parameters both to enforce a phase condition and to limit to odd frequency components, each optimization halving the number of trainable parameters. We start with a frequency cutoff of $K=16$ from the fixed point analysis, doubling the cutoff $K$ if the convergence condition of $\sqrt{ \ell } < 10^{-10}$ is not reached. The frequency cutoff rose to $K=1024$ for the longest converged orbits; converging an orbit took from seconds for $K=16$ to a few hours for $K=1024$, and memory requirements were primarily due to Hessian size.

\begin{figure}[ht]
  \centering
  \includegraphics[width=\columnwidth]{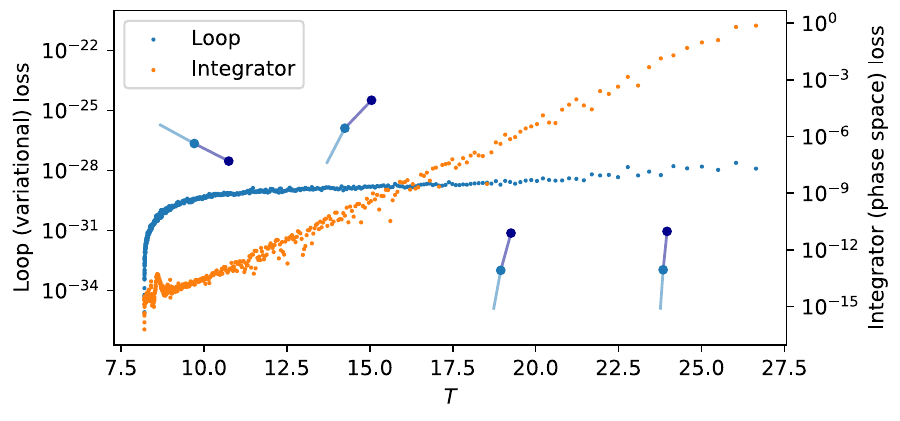}
  \caption{Loop and integrator losses for in-phase oscillations; variational convergence succeeds even for longer periods $T$ as masses approach vertical, while integrator error diverges.}
  \label{T-losses.figure}
\end{figure}

From small oscillations, we were able to successfully propagate all four orbit families to orbits approaching a period divergence, with masses within $5^{\circ}$ of vertical. Through the entire energy spectrum, loop loss remained below the convergence condition, while once-around phase space integrator error scaled exponentially with orbit period $T$ (Figure~\ref{T-losses.figure}). For orbits approaching vertical, the error after integrating once around diverges, despite the precision of the initial condition obtained from the loop. If using \texttt{scipy.integrate} at double precision, attempting to use an integrator method to converge longer period orbits will fail, as accumulating error from integration destroys the orbit. The variational method is robust to orbit period and chaotic phase space dynamics, demonstrating an advantage in finding orbits that are otherwise impossible to compute without extended-precision numerics.

\subsection{Bifurcation Detection}

To detect crossing orbit families, we look for additional zero eigenvalues along an orbit family, after removing the zeroes corresponding to time evolution and branch direction; we call these ``off-branch eigenvalues''. Plotting the minimum off-branch eigenvalue along an orbit family shows sharp drops to zero, indicating discrete crossing points as expected.

\begin{figure}[ht]
  \centering
  \includegraphics[width=\columnwidth]{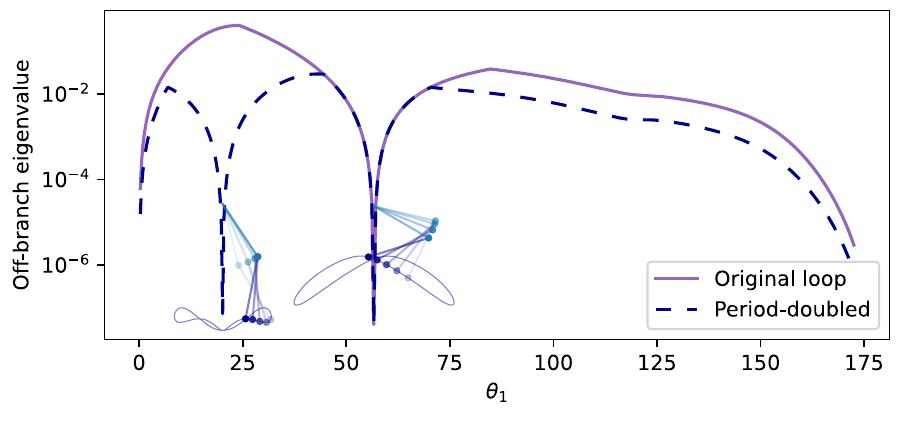}
  \caption{Minimum off-branch eigenvalue for out-of-phase oscillations; additional eigenvalue minimums give same period and period-doubling bifurcations leading to new orbit families.}
  \label{negative-min-eigenvalue.figure}
\end{figure}

To test for period-doubling bifurcations, we first generate a representation of a twice-wound loop by doubling the frequencies of all components and the period. We fill zeroes into the odd frequency components and take the Hessian, again extracting the minimum off-branch eigenvalue. In addition to the bifurcation points of the original loop, additional eigenvalue minimums corresponding to subharmonic perturbations are also seen (Figure~\ref{negative-min-eigenvalue.figure}).

Similar to fixed points, eigenvectors give loop perturbations that will maintain zero loss and a periodic orbit. As the dimensionality of the null subspace is greater, an orthogonalization is theoretically required to separate the perturbation subspace from the original orbit branch. In practice, a difference in numerical precision separated the orbit family and perturbation eigenvalues.

\subsection{Interpreting Orbits}

Eigenvector directions at bifurcation points lead to branching orbit families and additional connected orbits. By discovering continuous orbit families instead of individual disconnected orbits, we can classify orbits in terms of the families that they lie on, and the orbit families themselves can be classified based on their structure. %

\begin{figure}[ht]
  \centering
  \includegraphics[width=\columnwidth]{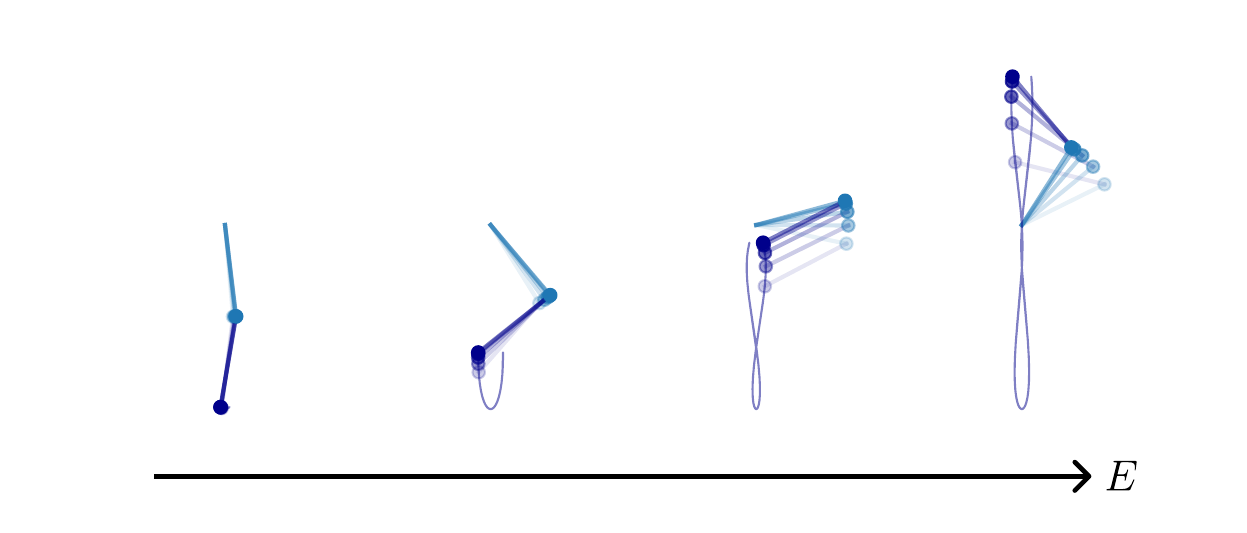}
  \caption{Out-of-phase oscillations at various energies along the continuous family; higher energy oscillations smoothly extend and deform the shapes of the low-energy normal mode.}
  \label{orbit-shape.figure}
\end{figure}

Along orbit families, higher energy periodic solutions can be understood as extensions of lower energy orbits. In the double pendulum, branches from fixed points start off with small oscillations, continuing to orbits where masses are very close to vertical. These higher energy periodic solutions can be interpreted as extensions of low-energy normal modes, and the shape of the oscillations also reflects this continuation (Figure~\ref{orbit-shape.figure}).

\begin{figure}[ht]
  \centering
  \includegraphics[width=\columnwidth]{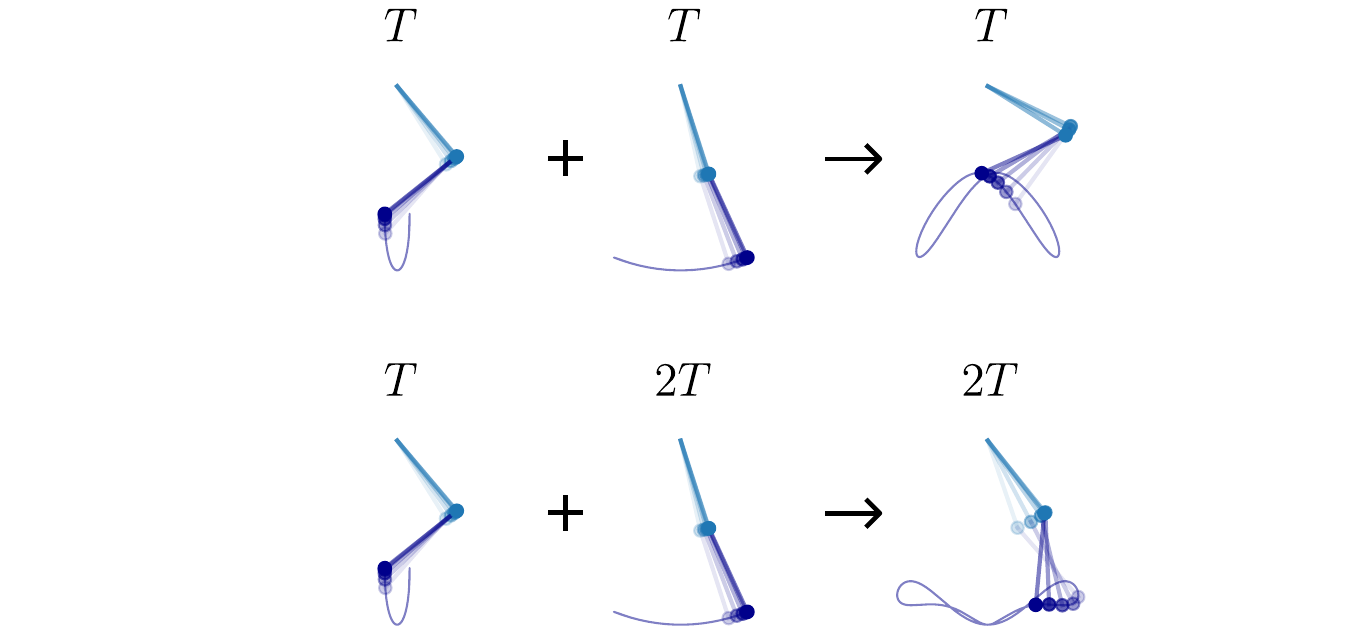}
  \caption{Bifurcating orbits are created by combinations of perturbations at same and higher multiples of period, which can be interpreted by following propagated branches.}
  \label{orbit-sums.figure}
\end{figure}

Through intersections of orbit families, we can also interpret motions that are not continuations of normal modes. If we consider a bifurcation point and the corresponding orbit, the eigenvector direction corresponds to a small perturbation leading to the branching orbit family. Other orbits on this branching family are continuations of this perturbation. We can understand this orbit family as a combination of the base orbit, and an oscillation perturbing from it (Figure~\ref{orbit-sums.figure}). In the case of period-doubling bifurcations, multiple oscillations of the base orbit may correspond to a single period of the perturbation.

By building up a collection of branching orbit families, we can trace a path to each orbit through the bifurcation structure, starting from fixed points and normal modes, then following orbit families and crossings. Motions can be systematically categorized in terms of base and branching families, and complicated motions can be intuitively broken down into sums of simpler oscillations.

\subsection{Extending Bifurcation Branches}

From each bifurcation point, we propagate orbits in the directions of the off-branch perturbations. We maintain the constraint on training dynamics, but relax the constraints on symmetry and phase.

When close to bifurcation points, unconstrained loss landscape geometry favors convergence towards the simpler orbit branch, instead of the bifurcation branch. The training dynamics constraint is especially important to guide initial convergence when starting to propagate a bifurcation, in order to maintain the direction of the initial perturbation and not jump back to the already known orbit family. %

\begin{figure}[ht]
  \centering
  \includegraphics[width=\columnwidth]{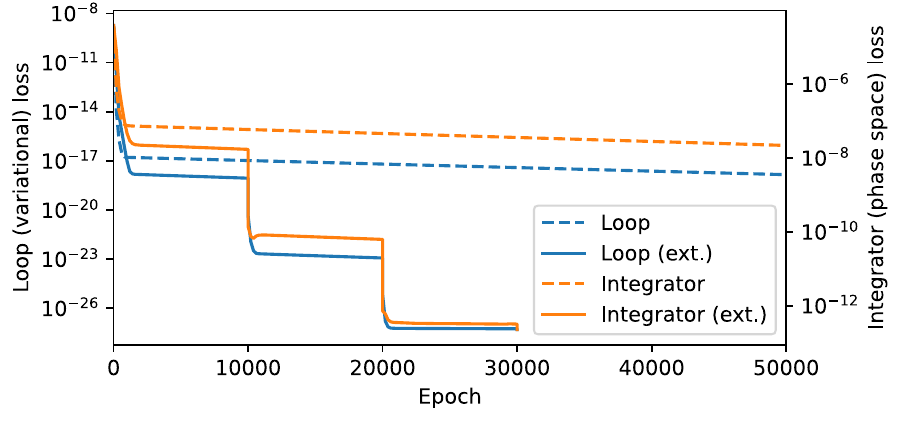}
  \caption{Comparison of loop and integrator losses during training with and without extrapolation; linear search greatly speeds up slow exponential convergence.}
  \label{extrapolate-comparison.figure}
\end{figure}

Due to the gradient signal being weaker in our constrained direction, final convergence is also notably slower in the vicinity of bifurcation points, despite an exponentially decreasing loss function. To accelerate a slow rate of exponential convergence, linear extrapolation was employed during the training process. The direction of convergence was estimated based on previous epochs and a linear search was conducted along the estimated direction, with the new initialization made at the training loss minimum along the line. Doing so sped up convergence significantly, with significant drops in losses seen at each extrapolation step (Figure~\ref{extrapolate-comparison.figure}).

\subsection{Phase Space Exploration}

Many examples of successfully extended branches were obtained across the phase space, with a majority of branches approaching period divergences or another termination condition. A variety of rich behaviors in the periodic orbit spectrum are revealed, such as an orbit family forming a closed loop and an additional bifurcation point. A plot of all converged periodic orbits is shown in Figure~\ref{full-plot.figure}, with branches labeled by base branch and type of bifurcation.

The structure of our converged branches and bifurcations matches orbit family theory. Both in-phase doublings show an orbit family passing continuously through the bifurcation point, with the first (ID1) looping back in on itself and the second (ID2) proceeding towards a period divergence. The first out-phase doubling (OD1) crosses the out-of-phase family (Out-Phase); the points on each side represent the same orbits (up to time reversal), but show initial conditions at different phases. The out-phase bifurcation branch (OB) continues past the bifurcation point to negative $\dot{\theta}_{2}$, following the same trace of initial angles but with opposite initial angular velocity.

The additional bifurcation point and branch (OD1B) is found along an out-of-phase period-doubling (OD1). Approaching a bifurcation point, the Hessian nullspace increases in dimension; this causes the computed zero-eigenvalue directions to vary depending on numerical precision. This bifurcation was found from two separate runs initially which followed the same orbit family before diverging.

The in-phase (IB, very short) and out-phase bifurcation branches (OB) cut off mid-propagation due to convergence failures. This is also likely due to a bifurcation point, but the convergence dynamics did not lead to a successful continued propagation along either orbit family. This failure mode is not present in the original branches, due to the phase and symmetry conditions. As the orbit branches are otherwise smooth, a constraint based on the recent orbit family direction could be used to continue propagation in future work.

We otherwise did not explicitly search for bifurcation points along non-symmetric orbit families. Theoretically, the periodic orbit structure is infinite; by continuing to compute off-branch eigenvalues and propagating new branches along perturbations, our method can be extended to search for additional orbit families up to numerical constraints.

\section{Conclusion}\label{conclusion.sec}

We have introduced a loss landscape perspective to searching for periodic orbits in dynamical systems. By probing loss curvatures using the Hessian, we can systematically explore the structure of orbit families, determining connected classes of orbits and bifurcations. We obtain a numerical continuation method in loop space, with variational initialization and convergence properties.

By combining the Rprop optimizer and linear extrapolation, we are able to converge orbits of varying complexity across the phase space of the ideal double pendulum. To our knowledge, this is the first survey of orbits that both discovers orbits off normal-mode branches and systematically classifies them into continuous families.

The success of this optimization routine also demonstrates the applicability of machine learning optimizers to problems in physical systems. While Rprop quickly gives satisfactory results without much tuning, taking advantage of different hyperparameters and algorithms may further improve performance. On an implementation level, constraints such as the propagation direction can be easily expressed using developed tools and methods from machine learning such as parametrizations.

A limitation of using the Rprop optimizer is that it is a gradient-based algorithm, which scales unfavorably in complexity for higher-dimensional systems compared to recent Jacobian-free methods. However, the Hessian probing of loss landscapes, which takes an orbit and finds eigenvector directions and new loop initializations, can be decoupled from methods to converge the initialization to a new orbit. Only a single Hessian evaluation is needed to determine the local loss landscape structure, and our propagation method is compatible with other variational methods in the literature. If computing the full Hessian is too expensive, approximations can be made by using reduced parameter loop representations or Hessian approximation algorithms.

Instead of finding individual, disconnected orbits, methodically searching orbits allows for solutions to be understood as part of a orbit family structure, and for an existing known structure to inform the search and discovery of new orbit candidates. By following branches on a bifurcation diagram, more complicated motions can be interpreted as combinations of simple oscillations, giving a more interpretable method for analyzing periodic solutions.

Promising future directions include both the continued application of machine learning methods to periodic orbits, and applying our method of systematically propagating through the orbit spectrum to a wide range of other dynamical systems.

\begin{acknowledgments}
  Z.L. and M.T. were supported by IAIFI through NSF grant PHY-2019786.
\end{acknowledgments}

\bibliographystyle{apsrev4-2}
\bibliography{Paper.bib}%

@article{Gelfreich_2002,
  title     = {Near strongly resonant periodic orbits in a Hamiltonian system},
  volume    = {99},
  issn      = {1091-6490},
  url       = {http://dx.doi.org/10.1073/pnas.212116699},
  doi       = {10.1073/pnas.212116699},
  number    = {22},
  journal   = {Proceedings of the National Academy of Sciences},
  publisher = {Proceedings of the National Academy of Sciences},
  author    = {Gelfreich, Vassili},
  year      = {2002},
  month     = oct,
  pages     = {13975–13979}
}

@inbook{Sbano2011,
  author    = {Sbano, Luca},
  editor    = {Meyers, Robert A.},
  title     = {Periodic Orbits of Hamiltonian Systems},
  booktitle = {Mathematics of Complexity and Dynamical Systems},
  year      = {2011},
  publisher = {Springer New York},
  address   = {New York, NY},
  pages     = {1212--1236},
  isbn      = {978-1-4614-1806-1},
  doi       = {10.1007/978-1-4614-1806-1_74},
  url       = {https://doi.org/10.1007/978-1-4614-1806-1_74}
}

@inbook{Gutzwiller1990,
  author    = {Gutzwiller, Martin C.},
  title     = {Periodic Orbits},
  booktitle = {Chaos in Classical and Quantum Mechanics},
  year      = {1990},
  publisher = {Springer New York},
  address   = {New York, NY},
  pages     = {75--86},
  isbn      = {978-1-4612-0983-6},
  doi       = {10.1007/978-1-4612-0983-6_7},
  url       = {https://doi.org/10.1007/978-1-4612-0983-6_7}
}

@article{MESTEL1987172,
    title = {Newton method for highly unstable orbits},
    journal = {Physica D: Nonlinear Phenomena},
    volume = {24},
    number = {1},
    pages = {172-178},
    year = {1987},
    issn = {0167-2789},
    doi = {https://doi.org/10.1016/0167-2789(87)90072-8},
    url = {https://www.sciencedirect.com/science/article/pii/0167278987900728},
    author = {Ben Mestel and Ian Percival}
}

@article{FARANTOS199591,
    title = {Methods for locating periodic orbits in highly unstable systems},
    journal = {Journal of Molecular Structure: THEOCHEM},
    volume = {341},
    number = {1},
    pages = {91-100},
    year = {1995},
    issn = {0166-1280},
    doi = {https://doi.org/10.1016/0166-1280(95)04206-L},
    url = {https://www.sciencedirect.com/science/article/pii/016612809504206L},
    author = {Stavros C. Farantos}
}

@Article{npg-14-615-2007,
    AUTHOR = {Saiki, Y.},
    TITLE = {Numerical detection of unstable periodic orbits in continuous-time dynamical systems with chaotic behaviors},
    JOURNAL = {Nonlinear Processes in Geophysics},
    VOLUME = {14},
    YEAR = {2007},
    NUMBER = {5},
    PAGES = {615--620},
    URL = {https://npg.copernicus.org/articles/14/615/2007/},
    DOI = {10.5194/npg-14-615-2007}
}

@article{PhysRevE.53.1206,
  title = {Method for computing long periodic orbits of dynamical systems},
  author = {Drossos, Lambros and Ragos, Omiros and Vrahatis, Michael N. and Bountis, Tassos},
  journal = {Phys. Rev. E},
  volume = {53},
  issue = {1},
  pages = {1206--1211},
  numpages = {0},
  year = {1996},
  month = {Jan},
  publisher = {American Physical Society},
  doi = {10.1103/PhysRevE.53.1206},
  url = {https://link.aps.org/doi/10.1103/PhysRevE.53.1206}
}

@article{PhysRevE.64.026214,
  title = {Detecting unstable periodic orbits in chaotic continuous-time dynamical systems},
  author = {Pingel, Detlef and Schmelcher, Peter and Diakonos, Fotis K.},
  journal = {Phys. Rev. E},
  volume = {64},
  issue = {2},
  pages = {026214},
  numpages = {10},
  year = {2001},
  month = {Jul},
  publisher = {American Physical Society},
  doi = {10.1103/PhysRevE.64.026214},
  url = {https://link.aps.org/doi/10.1103/PhysRevE.64.026214}
}

@article{Artuso_1990,
  title     = {Recycling of strange sets: I. Cycle expansions},
  volume    = {3},
  issn      = {1361-6544},
  url       = {http://dx.doi.org/10.1088/0951-7715/3/2/005},
  doi       = {10.1088/0951-7715/3/2/005},
  number    = {2},
  journal   = {Nonlinearity},
  publisher = {IOP Publishing},
  author    = {Artuso, R and Aurell, E and Cvitanovic, P},
  year      = {1990},
  month     = may,
  pages     = {325–359}
}

@article{Lan_2010,
  title     = {Cycle expansions: From maps to turbulence},
  volume    = {15},
  issn      = {1007-5704},
  url       = {http://dx.doi.org/10.1016/j.cnsns.2009.04.022},
  doi       = {10.1016/j.cnsns.2009.04.022},
  number    = {3},
  journal   = {Communications in Nonlinear Science and Numerical Simulation},
  publisher = {Elsevier BV},
  author    = {Lan, Y.},
  year      = {2010},
  month     = mar,
  pages     = {502–526}
}

@article{PhysRevE.102.052220,
  title = {Sensitivity of long periodic orbits of chaotic systems},
  author = {Lasagna, D.},
  journal = {Phys. Rev. E},
  volume = {102},
  issue = {5},
  pages = {052220},
  numpages = {15},
  year = {2020},
  month = {Nov},
  publisher = {American Physical Society},
  doi = {10.1103/PhysRevE.102.052220},
  url = {https://link.aps.org/doi/10.1103/PhysRevE.102.052220}
}

@article{S_NCHEZ_2010,
  title     = {ON THE MULTIPLE SHOOTING CONTINUATION OF PERIODIC ORBITS BY NEWTON–KRYLOV METHODS},
  volume    = {20},
  issn      = {1793-6551},
  url       = {http://dx.doi.org/10.1142/S0218127410025399},
  doi       = {10.1142/s0218127410025399},
  number    = {01},
  journal   = {International Journal of Bifurcation and Chaos},
  publisher = {World Scientific Pub Co Pte Lt},
  author    = {SÁNCHEZ, JUAN and NET, MARTA},
  year      = {2010},
  month     = jan,
  pages     = {43–61}
}

@article{Net_2015,
  title     = {Continuation of Bifurcations of Periodic Orbits for Large-Scale Systems},
  volume    = {14},
  issn      = {1536-0040},
  url       = {http://dx.doi.org/10.1137/140981010},
  doi       = {10.1137/140981010},
  number    = {2},
  journal   = {SIAM Journal on Applied Dynamical Systems},
  publisher = {Society for Industrial & Applied Mathematics (SIAM)},
  author    = {Net, M. and Sánchez, J.},
  year      = {2015},
  month     = jan,
  pages     = {674–698}
}

@article{Wulff_2006,
  title     = {Numerical Continuation of Symmetric Periodic Orbits},
  volume    = {5},
  issn      = {1536-0040},
  url       = {http://dx.doi.org/10.1137/050637170},
  doi       = {10.1137/050637170},
  number    = {3},
  journal   = {SIAM Journal on Applied Dynamical Systems},
  publisher = {Society for Industrial & Applied Mathematics (SIAM)},
  author    = {Wulff, Claudia and Schebesch, Andreas},
  year      = {2006},
  month     = jan,
  pages     = {435–475}
}

@book{Henon_1997,
  isbn      = {9783540638025},
  url       = {http://dx.doi.org/10.1007/3-540-69650-4},
  doi       = {10.1007/3-540-69650-4},
  journal   = {Lecture Notes in Physics Monographs},
  publisher = {Springer Berlin Heidelberg},
  year      = {1997},
  title     = {Generating Families in the Restricted Three-Body Problem},
  author    = {Michel Hénon}
}

@article{BARANGER198895,
    title = {The calculation of periodic trajectories},
    journal = {Annals of Physics},
    volume = {186},
    number = {1},
    pages = {95-110},
    year = {1988},
    issn = {0003-4916},
    doi = {https://doi.org/10.1016/S0003-4916(88)80018-6},
    url = {https://www.sciencedirect.com/science/article/pii/S0003491688800186},
    author = {M. Baranger and K.T.R. Davies and J.H. Mahoney}
}

@article{Wintner_1932,
  title     = {Grundlagen einer Genealogie der periodischen Bahnen im restringierten Dreikörperproblem: Erste Mitteilung Beweis des E. Strömgrenschen dynamischen Abschlußprinzips der periodischen Bahngruppen},
  volume    = {34},
  issn      = {1432-1823},
  url       = {http://dx.doi.org/10.1007/BF01180594},
  doi       = {10.1007/bf01180594},
  number    = {1},
  journal   = {Mathematische Zeitschrift},
  publisher = {Springer Science and Business Media LLC},
  author    = {Wintner, Aurel},
  year      = {1932},
  month     = dec,
  pages     = {321–349}
}

@article{Birkhoff1936,
  author    = {Birkhoff, George D.},
  journal   = {Annali della Scuola Normale Superiore di Pisa - Classe di Scienze},
  number    = {1},
  pages     = {9-50},
  publisher = {Scuola normale superiore},
  title     = {Sur le problème restreint des trois corps (second mémoire)},
  url       = {http://eudml.org/doc/82917},
  volume    = {5},
  year      = {1936}
}

@inproceedings{riedmiller1993direct,
  title        = {A direct adaptive method for faster backpropagation learning: The RPROP algorithm},
  author       = {Riedmiller, Martin and Braun, Heinrich},
  booktitle    = {IEEE international conference on neural networks},
  pages        = {586--591},
  year         = {1993},
  organization = {IEEE}
}

@incollection{NEURIPS2019_9015,
  title     = {PyTorch: An Imperative Style, High-Performance Deep Learning Library},
  author    = {Paszke, Adam and Gross, Sam and Massa, Francisco and Lerer, Adam and Bradbury, James and Chanan, Gregory and Killeen, Trevor and Lin, Zeming and Gimelshein, Natalia and Antiga, Luca and Desmaison, Alban and Kopf, Andreas and Yang, Edward and DeVito, Zachary and Raison, Martin and Tejani, Alykhan and Chilamkurthy, Sasank and Steiner, Benoit and Fang, Lu and Bai, Junjie and Chintala, Soumith},
  booktitle = {Advances in Neural Information Processing Systems 32},
  editor    = {H. Wallach and H. Larochelle and A. Beygelzimer and F. d\textquotesingle Alch\'{e}-Buc and E. Fox and R. Garnett},
  pages     = {8024--8035},
  year      = {2019},
  publisher = {Curran Associates, Inc.},
  url       = {http://papers.neurips.cc/paper/9015-pytorch-an-imperative-style-high-performance-deep-learning-library.pdf}
}

@article{PhysRevE.69.016217,
  title     = {Variational method for finding periodic orbits in a general flow},
  author    = {Lan, Yueheng and Cvitanovi\ifmmode \acute{c}\else \'{c}\fi{}, Predrag},
  journal   = {Phys. Rev. E},
  volume    = {69},
  issue     = {1},
  pages     = {016217},
  numpages  = {10},
  year      = {2004},
  month     = {Jan},
  publisher = {American Physical Society},
  doi       = {10.1103/PhysRevE.69.016217},
  url       = {https://link.aps.org/doi/10.1103/PhysRevE.69.016217}
}

@article{PhysRevE.98.042204,
  title     = {Accelerated variational approach for searching cycles},
  author    = {Wang, Ding and Wang, Peijie and Lan, Yueheng},
  journal   = {Phys. Rev. E},
  volume    = {98},
  issue     = {4},
  pages     = {042204},
  numpages  = {11},
  year      = {2018},
  month     = {Oct},
  publisher = {American Physical Society},
  doi       = {10.1103/PhysRevE.98.042204},
  url       = {https://link.aps.org/doi/10.1103/PhysRevE.98.042204}
}

@article{PhysRevE.105.014217,
  title     = {Constructing periodic orbits of high-dimensional chaotic systems by an adjoint-based variational method},
  author    = {Azimi, Sajjad and Ashtari, Omid and Schneider, Tobias M.},
  journal   = {Phys. Rev. E},
  volume    = {105},
  issue     = {1},
  pages     = {014217},
  numpages  = {14},
  year      = {2022},
  month     = {Jan},
  publisher = {American Physical Society},
  doi       = {10.1103/PhysRevE.105.014217},
  url       = {https://link.aps.org/doi/10.1103/PhysRevE.105.014217}
}

@article{Calv_o_2015,
  title     = {The double pendulum: a numerical study},
  volume    = {36},
  issn      = {1361-6404},
  url       = {http://dx.doi.org/10.1088/0143-0807/36/4/045018},
  doi       = {10.1088/0143-0807/36/4/045018},
  number    = {4},
  journal   = {European Journal of Physics},
  publisher = {IOP Publishing},
  author    = {Calvão, A M and Penna, T J P},
  year      = {2015},
  month     = may,
  pages     = {045018}
}

@article{Jim_nez_L_pez_2024,
  title     = {Chaos and Regularity in the Double Pendulum with Lagrangian Descriptors},
  volume    = {34},
  issn      = {1793-6551},
  url       = {http://dx.doi.org/10.1142/S0218127424502018},
  doi       = {10.1142/s0218127424502018},
  number    = {16},
  journal   = {International Journal of Bifurcation and Chaos},
  publisher = {World Scientific Pub Co Pte Ltd},
  author    = {Jiménez-López, Javier and García-Garrido, Víctor J.},
  year      = {2024},
  month     = nov
}

@article{PhysRevLett.126.180604,
  title     = {Machine Learning Conservation Laws from Trajectories},
  author    = {Liu, Ziming and Tegmark, Max},
  journal   = {Phys. Rev. Lett.},
  volume    = {126},
  issue     = {18},
  pages     = {180604},
  numpages  = {6},
  year      = {2021},
  month     = {May},
  publisher = {American Physical Society},
  doi       = {10.1103/PhysRevLett.126.180604},
  url       = {https://link.aps.org/doi/10.1103/PhysRevLett.126.180604}
}

@article{Kaheman_2023,
  title     = {Saddle transport and chaos in the double pendulum},
  volume    = {111},
  issn      = {1573-269X},
  url       = {http://dx.doi.org/10.1007/s11071-023-08232-0},
  doi       = {10.1007/s11071-023-08232-0},
  number    = {8},
  journal   = {Nonlinear Dynamics},
  publisher = {Springer Science and Business Media LLC},
  author    = {Kaheman, Kadierdan and Bramburger, Jason J. and Kutz, J. Nathan and Brunton, Steven L.},
  year      = {2023},
  month     = jan,
  pages     = {7199–7233}
}

@article{Yu_1998,
  title     = {ANALYSIS OF NON-LINEAR DYNAMICS AND BIFURCATIONS OF A DOUBLE PENDULUM},
  volume    = {217},
  issn      = {0022-460X},
  url       = {http://dx.doi.org/10.1006/jsvi.1998.1781},
  doi       = {10.1006/jsvi.1998.1781},
  number    = {4},
  journal   = {Journal of Sound and Vibration},
  publisher = {Elsevier BV},
  author    = {Yu, P. and Bi, Q.},
  year      = {1998},
  month     = nov,
  pages     = {691–736}
}

@article{Yagasaki_1996,
  title     = {A simple feedback control system: Bifurcations of periodic orbits and chaos},
  volume    = {9},
  issn      = {1573-269X},
  url       = {http://dx.doi.org/10.1007/BF01833363},
  doi       = {10.1007/bf01833363},
  number    = {4},
  journal   = {Nonlinear Dynamics},
  publisher = {Springer Science and Business Media LLC},
  author    = {Yagasaki, K.},
  year      = {1996},
  month     = apr,
  pages     = {391–417}
}

@article{Liu_2023,
  title     = {Non-quantum chirality and periodic islands in the driven double pendulum system},
  volume    = {177},
  issn      = {0960-0779},
  url       = {http://dx.doi.org/10.1016/j.chaos.2023.114254},
  doi       = {10.1016/j.chaos.2023.114254},
  journal   = {Chaos, Solitons \& Fractals},
  publisher = {Elsevier BV},
  author    = {Liu, Zeyi and Rao, Xiaobo and Gao, Jianshe and Ding, Shunliang},
  year      = {2023},
  month     = dec,
  pages     = {114254}
}

@article{Williams_2023,
  title     = {A compound double pendulum with friction},
  volume    = {10},
  issn      = {2666-3597},
  url       = {http://dx.doi.org/10.1016/j.finmec.2022.100164},
  doi       = {10.1016/j.finmec.2022.100164},
  journal   = {Forces in Mechanics},
  publisher = {Elsevier BV},
  author    = {Williams, Hollis},
  year      = {2023},
  month     = feb,
  pages     = {100164}
}

\end{document}